\pdfoutput=1

\documentclass[11pt]{article}

\usepackage[preprint]{acl}

\usepackage{times}
\usepackage{latexsym}

\usepackage[T1]{fontenc}

\usepackage[utf8]{inputenc}

\usepackage{microtype}

\usepackage{inconsolata}

\usepackage{graphicx}

\usepackage{amsfonts}
\usepackage{booktabs}
\usepackage{makecell} 
\usepackage{multirow}
\usepackage{multicol}
\usepackage{enumitem}
\usepackage{amsmath}
\usepackage{algorithm} 
\usepackage{algorithmic}

\title{Random Forest-of-Thoughts: Uncertainty-aware Reasoning for Computational Social Science}

\author{Xiaohua Wu$^{1,2}$, Xiaohui Tao$^{3}$, Wenjie Wu$^{2}$, Yuefeng Li$^{1}$, {Lin Li$^{2}$}\thanks{Corresponding Author} \\
  $^{1}$Queensland University of Technology, Brisbane 4000, Australia \\
  $^{2}$Wuhan University of Technology, Wuhan 430070, China \\
  $^{3}$University of Southern Queensland, Springfield 4300, Australia\\
  \texttt{\{xhwu,cathylilin\}@whut.edu.cn} \quad
  \texttt{xiaohui.tao@unisq.edu.au} \quad
  \texttt{y2.li@qut.edu.au} \quad
  }

\begin{document}
\maketitle
\begin{abstract}

Social surveys in computational social science are well-designed by elaborate domain theories that can effectively reflect the interviewee's deep thoughts without concealing their true feelings. The candidate questionnaire options heavily depend on the interviewee's previous answers, increasing the complexity of social survey analysis, as well as the time and expertise required. The ability of large language models (LLMs) to perform complex reasoning is well-enhanced by prompting learning such as Chain-of-thought (CoT) but still confined to left-to-right decision-making processes or limited paths during inference. This means they can fall short in problems that require exploration and uncertainty searching. In response, a novel large language model prompting method, called Random Forest of Thoughts (RFoT), is proposed for generating uncertainty reasoning to fit the area of computational social science. The RFoT allows LLMs to perform deliberate decision-making by generating diverse thought space and randomly selecting the sub-thoughts to build the forest of thoughts. It can extend the exploration and prediction of overall performance, benefiting from the extensive research space of response. The method is applied to optimize computational social science analysis on two datasets covering a spectrum of social survey analysis problems. Our experiments show that RFoT significantly enhances language models' abilities on two novel social survey analysis problems requiring non-trivial reasoning.

\end{abstract}

\section{Introduction}
Social surveys form the foundation of contemporary empirical studies in social science and computational science \cite{computational_personality_survey_2022,AIforSocialScience2024}. With the rise of empiricism in the mid to late 19th century, the social survey became a primary method for understanding society, providing valuable insights into various aspects such as demographics, employment \cite{HappyDB_2018}, health, and education \cite{nardi_survey_research_2018}. The social survey typically employs a systematic sampling approach to generate a probability sample, combined with carefully designed questionnaires, to ensure analysis accuracy. The sample's representativeness is crucial, enabling statistical generalization to the target population \cite{nature_2021}. Based on the Ecological Momentary Assessment (EMA) \cite{EMA_2021}, leveraging artificial intelligence technology to rapidly, objectively, and automatically conduct computational social science analysis has become a new trend. Conducting high-quality, large-scale social survey analyses has become increasingly challenging due to the time and expertise required.

With the development of large language models (LLMs), previous works try numerous attempts based on LLMs aimed at social survey analysis including questionnaire generation \cite{ChatGPT_vs_Social_Surveys_2024}, understanding public opinion \cite{argyle_2023}, and the alignment of the distribution patterns of responses from the ``homo silicus” and human respondents \cite{ChatGPT_vs_Social_Surveys_2024}.

\begin{figure}[ht]
    \centering
    \includegraphics[width=1\linewidth]{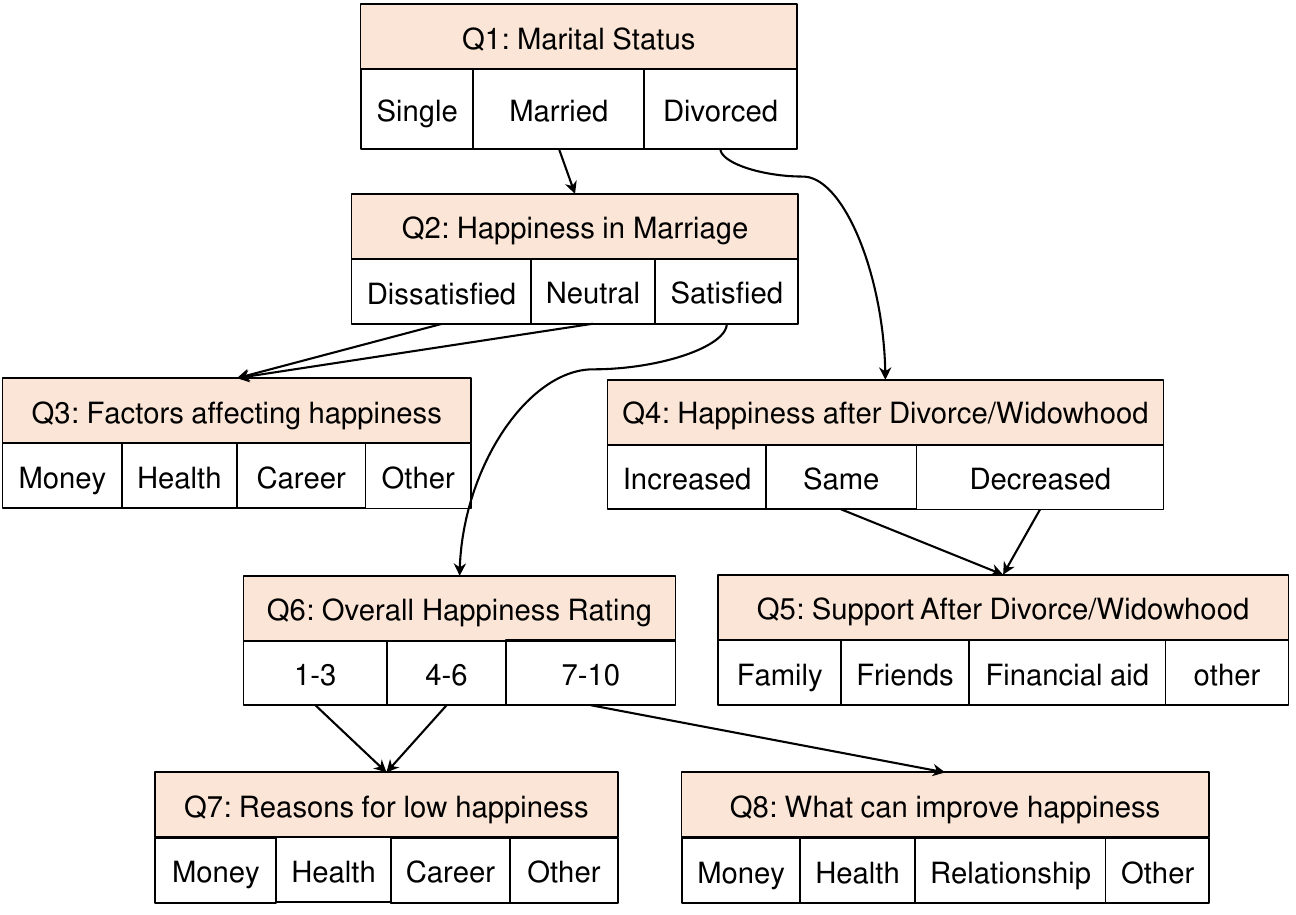}
    \caption{The questionnaire analysis with uncertainty question-answer pairs. }
    \label{fig:Random_Questionnaire}
\end{figure}

The previous works usually utilized DNN-based models such as LSTM, Transformer, and pertaining and fine-tuning models to assess the psychological states. Recently, large language models (LLMs) have been introduced to provide more stronger performance for the completeness of psychological state analysis. To further improve the model performance, a chain of thought prompting is proposed to enhance the reasoning response with a series of intermediate natural language reasoning steps \cite{Chain_of_Thought_2022,Automatic_Chain_of_Thought_2023,Active_Prompting_2024}. However, these models only consider the questionnaires as text, ignoring the randomness of the questions depending on respondents and domain theories behind the questionnaire designed in social science. As shown in Fig. \ref{fig:Random_Questionnaire}, the questionnaire has different lines based on the previous answer. For example, question four should not be answered if the respondent is married, and question seven needn't be answered if the overall happiness rating is under six. The random forest (RF) can classify the labels by combining various features according to their contribution, which has been widely used in various intelligent decisions. Inspired by some social science theories, the Random Forest of Thoughts (RFoT) is proposed to explore the randomness and model domain theories to construct various thought chains for exploring rich solution spaces. Specifically, the Chain of Thought-based LLMs \cite{Chain_of_Thought_2022} is employed to generate stable and high-quality responses based on the questionnaire input. The key thoughts are extracted from the response by a novel strategy involving a model-agnostic contribution evaluation method. 

Our pivotal contributions encapsulate:

\begin{enumerate}
    \item We propose a novel method named random forest of thoughts (RFoT) to search from a rich thoughts space and generate a more significant reasoning solution for better computational social science analysis performance.
    \item Domain theories reflected from different aspects behind questionnaires are considered and utilized to find more trustworthy reasoning steps, which can be as explanations for RFoT outputs. 
    \item Proposed iterative chain of thought prompting iteratively generates the thoughts from each aspect with a greedy strategy for rich alternative thoughts.
    \item The experimental results based on two popular survey datasets extensively demonstrate the superiority and effectiveness of the proposed method RFoT achieved up to 78.43\% success rate and 80.52\% weighted-f1 scores.
\end{enumerate}

\section{Related Work}

\subsection{Computational Social Science Analysis}
The current methods in computational social science require an amount of time to assess the labels required by professionals. For example, previous research has identified that mental state is primarily associated with factors such as income, health, family, and others, through regression analysis and machine learning (ML) \cite{Saputri_2015,Yu_2017,Laaksonen_2018}, which is highly interpretability. With the advancement of deep neural networks (DNN), several researchers have proposed various DNN-based methods to analyze the relationships between factors and mental state \cite{Weizhao2019,ijcai2022_Lilin}. Based on social science theories, how to use artificial intelligence technology to quickly, objectively, and automatically complete mental state assessments has become a new trend. A follow-up study \cite{TCSS_2025,ADMA2024} utilized DNN-based methods to consider the domain knowledge in the designed questionnaires such as information on aspects but not random question-answer pairs due to different respondents. 

\subsection{Prompt Learning}


\paragraph{Pretraining and Fine-tuning}
Recently years have witnessed a rapid development of large language models (LLMs), which have a strong ability in many language-understanding tasks. The heavy computational burden and limited universality largely restrict the application of LLMs in various problems. For the first issue, there are two lines of research including parameter-efficient fine-tuning (PEFT) \cite{Parameter_Efficien_2022} and parameter quantization \cite{frantar2022optq,xiao2023smoothquant}. One of the most popular approaches is the low-rank adaptation (LoRA) \cite{LoRA_2022}, which is an improved fine-tuning method where instead of fine-tuning all the weights that constitute the weight matrix of the pre-trained large language models. Recently, joint adaptation and quantization methods have been proposed for achieving the objectives of both parameter-efficient adaptation and computation-efficient tuning and deployment \cite{QLoRA_2023,QA_LoRA_2024}. It can further improve the efficiency and scalability of LLMs as well as mitigate the negative impact of quantization errors. 

\paragraph{Input-output (IO) Prompting}

Zero-shot and few-shot prompting \cite{Zero_Shot_2022} are the most represented input-output prompting methods, which provide natural language instructions that describe the task and specify the expected output. This approach enables the LLMs to construct a context that refines the inference space, yielding a more accurate output \cite{Zero_Shot_Prompting_2024}. Numerous studies \cite{Few_Shot_2020,MEGA_2023} demonstrated that few-shot learning offers superior performance when compared to zero-shot learning setup. These works demonstrate that various prompting methods benefit the LLMs.

\paragraph{X-of-Thought Prompting}

For providing a way to solve complex problems that are not easily formalized, chain-of-thought (CoT) prompting \cite{Chain_of_Thought_2022} was proposed to address cases where the mapping of question $x$ to answer $y$ is non-trivial. It enhances the performance of LLMs by improving their reasoning capabilities, allowing them to solve complex tasks through step-by-step thinking. Recently, CoT has been utilized in various research areas such as question-answering systems \cite{QA_LLM_2023}, mathematical reasoning \cite{multilingual_CoT_2023}, and mini crosswords and creative writing \cite{Self_Consistency_COT_2023,Tree_of_thoughts_2023}. However, these models can not fit the questionnaire analysis in computational social science and do not consider the answer differences between respondents. 

Uncertainties in intermediate decision points are beneficial to combining various features according to the importance of features and have been widely used in various intelligent decisions \cite{TouT_2024}. Inspired by this, we employ LLMs to generate various linguistic thoughts to prompt LLMs reasoning strategy from multiple levels including key feature triggers, multi-aspect analysis, and overall assessment. The thoughts are evaluated by a post-hoc explanation method and then randomly joined in the thought subset for the next stage of reasoning.

\section{Problem Statement}

Let $\mathcal{D} = \{ (c_i, q_i, a_i) \} ^m_{i=1}$ denote a dataset consisting of $m$ multi-turn interviews. $c_i$ represents a category label describing the thematic classification of this dialogue turn. $q_i$ is a question posed in this turn questionnaire. $a_i$ represents the corresponding response provided by respondents. The goal of this work is to extract multi-level semantic features—referred to as thoughts $\mathcal{T}_i = [\mathcal{T}_i^1; \mathcal{T}_i^2; \dots; \mathcal{T}_i^n]$ — from the dataset using LLMs. Let $y_i \in \mathcal{Y}$ denote the mental state corresponding to each questionnaire turn. The objective is to learn a mapping $f:\mathcal{T} \to y$, parameterized by a predictive model $f_{\theta}$ such that:

\begin{equation}
    y_i = f_{\theta} (x; <\mathcal{T}_i>)
\end{equation}

A random forest-inspired prompting is employed to aggregate predictions from the $M$ chain of thought promptings, each trained on a random thought subset from the candidates for rich exploring space. The predicting function is given by: 

\begin{equation}
    y_i = \frac{1}{M} \sum_{m=1}^{M} f_\theta^{(m)}(x;<\mathcal{T}_i>)
\end{equation}

\begin{figure*}[ht]
    \centering
    \includegraphics[width=1\linewidth]{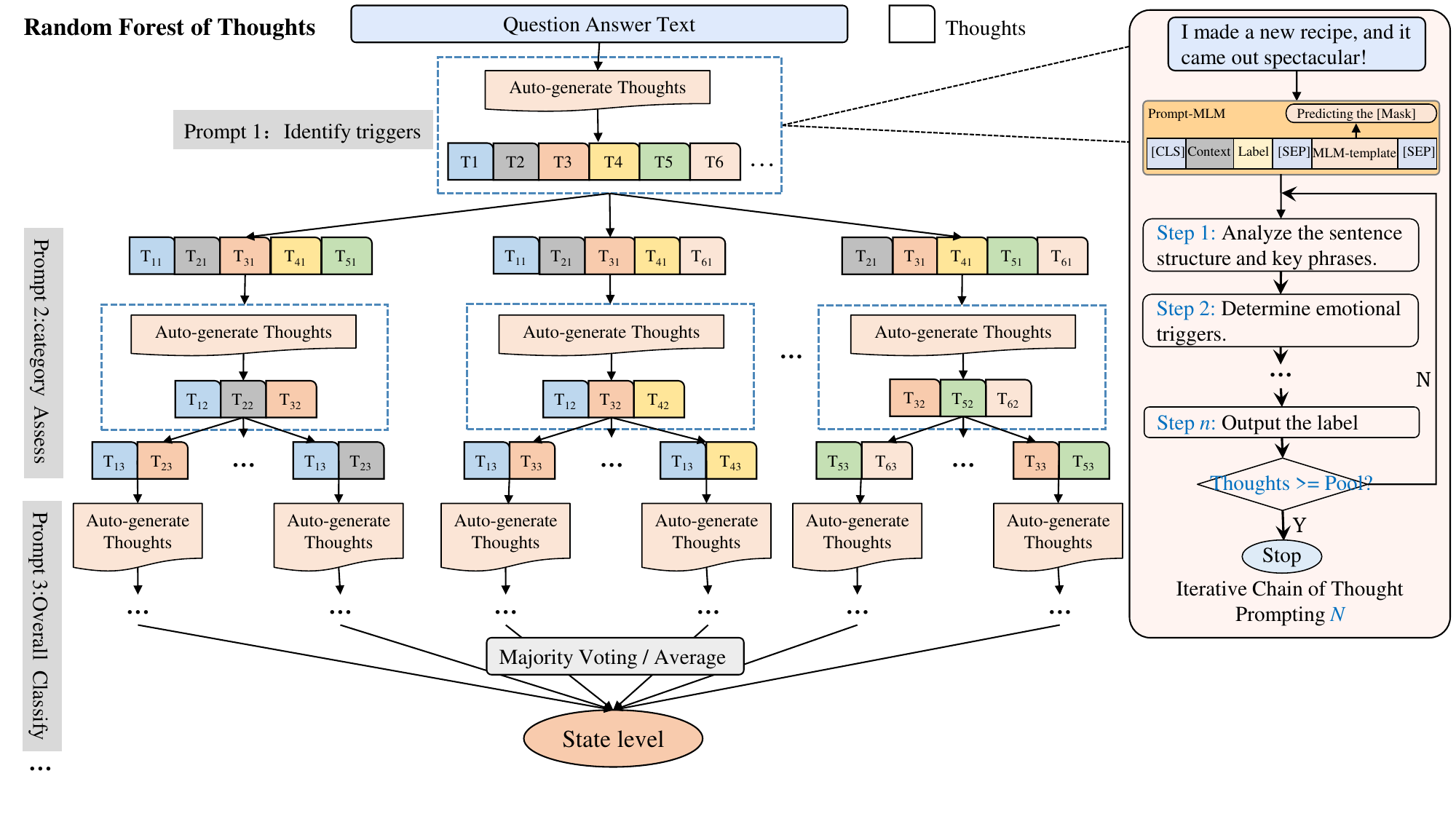}
    \caption{The framework of our proposed RFoT. Text input is decomposed into intermediate steps that are reconstructed as a prompt step $N$ boxed by a blue line. Each prompt step is an Iterative Chain of Thought (ICoT) as shown on the right.}
    \label{fig:RFOT_Framework}
\end{figure*}

\section{Methodology}

Uncertainties exploration is used in various classification and regression tasks, which can get a strong performance via random feature combinations to search for the best one from a huge result space. Inspired by that, we proposed a novel LLMs prompting method framework named Random Forest of Thoughts (RFoT), shown in Fig.~\ref{fig:RFOT_Framework}. Prompt learning is employed to translate the mental health posts into linguistic cues to decompose the intermediate steps. The iterative chain of thought (ICoT) is proposed for an iterative model generating the thoughts for each category. Instead of information entropy, value each state or vote strategy \cite{Tree_of_thoughts_2023} for evaluating the contribution of each feature, the thoughts are evaluated by a post-hoc explanation method with a theoretical and fundamental basis in this work \cite{Shapley_2016}. The RFoT is designed as an ensemble learning for exploring the best strategies from the rich thoughts space.

\subsection{Linguistic Cues Construction}

\begin{figure}[ht]
    \centering
    \includegraphics[width=1\linewidth]{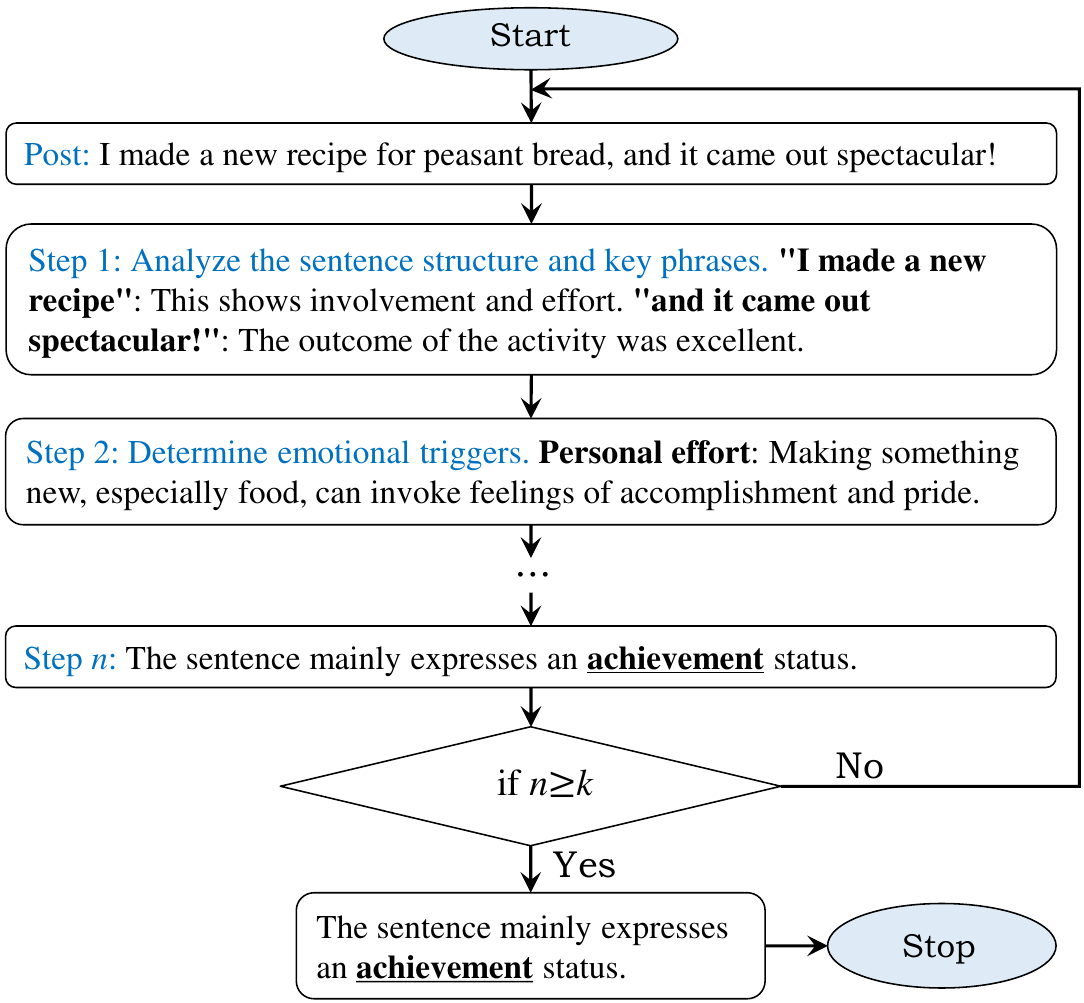}
    \caption{The thoughts generation from an aspect by the proposed ICoT.}
    \label{fig:Prompt_step}
\end{figure}

Linguistic cues refer to specific patterns, words, phrases, or structures in language that provide insights into a speaker's emotions, intentions, or attitudes. Analyzing these cues in a social computing context can help predict emotions, such as happiness, sadness, and depression, based on text input. Linguistic Cues are beneficial to language models for feature extraction, we therefore construct the mental health post into some linguistic cues as intermediate prompt steps. Many previous studies in a variety of areas have used Linguistic Inquiry and Word Count (LIWC) for obtaining linguistic cues and training machine learning classifiers in text classification and predictive outcome analysis \cite{MCHANEY2018,Hanks2022}. Recently, prompt learning is a novel method that can fully utilize linguistic cues for context representation and modeling. The linguistic cues are represented by some intermediate steps such as in Fig.~\ref{fig:Prompt_step}.

\subsection{Candidate Thoughts Generation Strategy}

The mental state is affected by different aspects based on the social science theory. For example, \citet{Bognar2010authentic} claims that the mental state is mainly affected by five aspects including positive emotion, engagement, relationships, meaning and purpose, and accomplishment. Inspired by these theories, the ICoT is proposed to generate thoughts for different aspects.

Given a questionnaire with $m$ turns question-answer pairs labeled by different categories as training data $\mathcal{D} = \{(q_1, a_1), (q_2, a_2), \dots, (q_m, a_m) \}$ with each $(q_i, a_i)$ indicating the question-answer pair, the goal of ICoT is to annotate these questions by constructing a new exemplar set $E = \{ (q_1, c_1, a_1), (q_2, c_2, a_2), \dots, (q_n, c_n, a_n) \}$ with a series of reasoning steps $c$ to reasoning the correct sub-category mental state out. 

\paragraph{Iterative Chain of Thought}
Previous works have shown that different layers of a language model encode different linguistic information within a sentence \cite{AkenWLG19}. Inspired by that, various language models with different layers are component learners and are integrated into a forest to achieve better prediction performance. For each category, we utilize the LLMs with Chain of Thought to generate the thoughts from multi-level aspects. It can be formulated as a multi-level hierarchical function:
\begin{equation}
    \mathcal{F}_{LLM} = (\mathcal{Q, A, C}) \to (A, K, R),
\end{equation}
where the $\mathcal{Q}$ represents a set of questions,  $\mathcal{A}$ is the corresponding answers, and $\mathcal{C}$ denotes different categories. Given a question-answer pair $(q_i, a_i)$, we define the extraction function of $L_i$th level thoughts:
\begin{equation}
    \begin{split}
        \mathcal{T'}_{L_1} & = A(q_i, a_i) = \mathcal{F}_{LLM}^{(L_1)}(q_i, a_i) \\
        \mathcal{T'}_{L_2} & = K(q_i, a_i) = \mathcal{F}_{LLM}^{(L_2)}A(q_i, a_i) \\
        \mathcal{T'}_{L_3} & = R(q_i, a_i) = \mathcal{F}_{LLM}^{(L_3)}K(q_i, a_i)
        \label{multi_level_thoughts}
    \end{split}
\end{equation}
where $\mathcal{T}_{L_i}$ is the thoughts of level $L_i$ generated by LLMs. $A$, $K$, and $R$ represent the aspects, emotional keywords, and final responses from LLMs, respectively. The thoughts generated at each level are the candidates of our proposed random forest of thoughts.

\subsection{Random Forest of Thoughts}

The candidate thoughts are generated by ICoT in the previous section. Selecting the most significant thoughts to be considered in RFoT is a key issue. 

\paragraph{Explanation-based Thoughts Evaluation}

Inspired by explainable artificial intelligence (XAI), we proposed a novel thought evaluation method to select the $k$ most important thoughts generated by ICoT using the post-hoc and model-agnostic explanation methods. It can reduce the noises contained in the questionnaires and improve the computing effectiveness. Compared to other post-hoc explanation methods like LIME \cite{Ribeiro2016} and DeepLIFT \cite{Shrikumar_2017}, Shapley values uniquely satisfy the accuracy, missingness, and consistency properties of feature attribution \cite{Shapley_2016}. This allows for more trustworthy and theoretical guarantees for the quantitative evaluation. 
We consider mental state prediction as a cooperative task with the ultimate goal of accurately predicting the label with all present thoughts. Therefore, the importance of thought is represented by $\phi_{j}$ which is defined as follows:
 \begin{equation}
    \begin{split}
 	\phi_{j}(v) =& \textstyle \sum_{S\subseteq J \setminus \left \{ x_j \right \}}\frac{|S|!(|J|-|S|-1)!}{|J|!} \\
                & \left[v(S\cup  \left \{ x_j \right \}) - v(S) \right],     \label{Shapley}
    \end{split}
 \end{equation}
where $j=1,...,|J|$, $J = \left \{ x_1,\cdots, x_{|J|} \right \}$ and $S\subseteq J \setminus \left \{ x_j \right \} $ denotes all possible subsets of thought set $J$, which excludes the thought $j$ and consists of $|S|$ thoughts. $|S|!(J-|S|-1)!/J!$ is the possibility of a subset $S$. $v(S\cup \left \{ x_j \right \})- v(S)$ means the marginal contribution of thought $j$ where $v(x)\in \mathbb{R}$ denotes the model output when thought in $S$ is present. In other words, the gain is a weighted average over contribution function difference in all subsets $S$, excluding the thought $j$.

\paragraph{Candidate Thoughts Generation}

we select a subset including top-$k$ thoughts with the highest scores $\phi_{j}(v)$ and relative aspects, which is shown in Equ. \ref{top_k_thoughts}.
\begin{equation}
    \mathcal{T}^* = \arg\max_{\mathcal{T}' \subseteq \mathcal{T}} \sum_{T_j \in \mathcal{T}'} \phi_{j}(v), \quad \text{subject to } |\mathcal{T}'| = k,
    \label{top_k_thoughts}
\end{equation}
where the $\mathcal{T}$ represents the thoughts generated by ICoT, $\mathcal{T}^*$ is the subset including top-$k$ important thoughts.

\paragraph{RFoT Construction Strategy}

A crucial challenge in predicting mental states through questionnaires is how to model and represent the uncertainty in responses from different individuals. Instead of selecting a fixed deterministic set of thoughts like the Forest of Thoughts \cite{Forest_of_Thought_2024}, Tree of Thoughts \cite{Tree_of_thoughts_2023}, and Self-consistency Chain of thought \cite{Self_Consistency_COT_2023}, we combine bootstrap sampling, a randomized sampling method, for constructing RFoT. Bootstrap sampling \cite{Bootstrap_Sampling_2022} is used in a machine learning ensemble algorithm called bagging. It helps in avoiding over-fitting and improves the stability of machine learning algorithms. 
Bootstrap sampling randomly and equally samples the original question-answer pairs in a questionnaire to construct many approximate independent identically distributed (i.i.d) new samples of the original questionnaire, which can comprehensively consider the situations of different respondents. As defined by Equ.~\ref{Bootstrap_sampling}, the probability of selecting a thought is proportional to its importance score $\phi_{\mathcal{T}_{L_j}}(v)$.
\begin{equation}
    P(\mathcal{T}_{L_j} \in \mathcal{T^*}) = \frac{\phi_{\mathcal{T}_{L_j}}(v)}{\sum_{j=1}^{n} \phi_{\mathcal{T}_{L_j}}(v)},
    \label{Bootstrap_sampling}
\end{equation}
From the selected subset $\mathcal{T}_{L_j}$, we construct multiple trees $T_1, T_2, \dots, T_m$ using randomization. Specifically, we first select the multiple root nodes $R=\{ T_{r_1}, T_{r_1}, \dots, T_{r_m} \}$ based on Equ.~\ref{root_selection} with high contribution of thoughts.
\begin{equation}
    P(T_{r_j}~is~root) = \frac{\phi_{j}(v)}{\sum_{T_{r_j} \in \mathcal{T}^*} \phi_{j}(v)}
    \label{root_selection}
\end{equation}

\begin{algorithm}[tb]
\caption{RFoT construction by DFS}
\label{alg:DFS}
    \textbf{Require}: tree number $m$, data $\mathcal{D}$, thoughts $\mathcal{T}'$ generated by ICoT;
    \begin{algorithmic}[1] 
        \STATE Drawing question-answer pairs subset $\mathcal{X}$ with replacement from $\mathcal{D}$;
        \FOR {$x$ in $\mathcal{X}$}
            \STATE Generating thoughts $\mathcal{T}'$ by ICoT;
            \STATE Drawing subset thoughts $\mathcal{T}^*$ without replacement from $\mathcal{T}'$;
            \FOR {$\mathcal{T}_{L_j}$ in $\mathcal{T}^*$}
            \STATE Selecting the root of each tree by Equ. \ref{root_selection};
            \STATE Finding the best-split thought;
            \STATE Building the left sub-tree through DFS($\mathcal{T}^{left}_i$);
            \STATE Building the right sub-tree through DFS($\mathcal{T}^{right}_i$);
            \ENDFOR
        \STATE Appending thought trees to the random forest of thoughts;
        \ENDFOR
        \STATE \textbf{Return} Random Forest of Thoughts
    \end{algorithmic}
\end{algorithm}

Depending on the forest structure, one can plug and play different search algorithms. We explore one relatively simple and effective tree expansion algorithm and leave more advanced ones in future work. Depth-first search (DFS) is preferable to Breadth-First search (BFS) in certain computational scenarios due to its lower memory consumption, efficiency in deep search spaces, and suitability for recursive problem-solving \cite{DFS_2024}. The RFoT construction by depth-first search (DFS) is presented in Algorithm \ref{alg:DFS} for a clearer calculation process of RFoT construction. Step 1 samples the subset of thoughts $\mathcal{T}^*$ from $\mathcal{T}'$ generated by ICoT. Steps 2-8 run in a loop to construct the random forest of thoughts. Specifically, step 3 draws $\mathcal{T}_{L_j}$ without replacement from $\mathcal{T}^*$. Step 4 selects the root of each tree by Equ. \ref{root_selection}, step 5 is to find a best-split thought through information gain rate, and steps 6-7 are for building the left and right sub-tree. Step 9 appends all trees to the random forest of thoughts.

\section{Experiment}

\paragraph{Experiment Setup}
In all benchmarks, 
Llama3-8B\footnote{https://ollama.com/library/llama3} \cite{meta_llama3}, and Qwen2.5-7B\footnote{https://ollama.com/library/qwen2.5:7b} 
are used as the base LLMs to verify their generalization because of the open-source policy. We compare the performance by different prompting strategies including standard zero-shot IO prompting \cite{Zero_Shot_2022}, Fine-tuning such as Low-Rank Adaptation (LoRA) \cite{LoRA_2022}, chain-of-thought (CoT) prompting \cite{Chain_of_Thought_2022}, Self-consistency CoT prompting (SC-CoT) \cite{Self_Consistency_COT_2023}, and tree-of-thoughts (ToT) prompting \cite{Tree_of_thoughts_2023}. As in previous works, we run each prompt for 100
samples and average the results for analysis.

\paragraph{Metrics}
The success rate \cite{Tree_of_thoughts_2023,Chain_of_Thought_2022} is used to evaluate LLMs' reasoning performance. In addition, the generation consistency or format adherence of LLMs is employed to evaluate whether the response is consistent with the predefined prompt format \cite{wangyue_2024}. It is defined as Equ.~\ref{consistency}.
 \begin{equation}
    C_F(G, E) = 1 - \frac{\text{Dist}(G(P), E(P))}{\max(|G(P)|, |E(P)|)}
    \label{consistency}
 \end{equation}
where $G(P)$ represents the generated response from the LLM given the prompt $P$, $E(P)$ denotes the predefined format or structure. $|G(P)|$ and $|E(P)|$ are the lengths of $G(P)$ and $E(P)$, respectively. Finally, the weighted F1 and running time per sample are reported for a more comprehensive analysis.

\paragraph{Environment} The experiments are conducted in Python 3.12 and PyTorch 2.3.0, running on a standard server equipped with an RTX 3080x2 20G GPU and an Intel(R) Xeon(R) Platinum 8352V CPU @ 2.10GHz. 

\subsection{Results and Discussion}
Mental state prediction based on questionnaires is a popular problem in social science, such as happiness prediction. In this work, we conduct our proposed LLMs prompting method RFoT on this problem with two datasets to explore the performance.

\begin{table*}
\footnotesize
\centering
\tabcolsep=0.8mm
\caption{Reasoning results on two datasets by two LLMs}
\label{tab:LLMs_Happiness_results}
\begin{tabular}{lclcccc}
\toprule
Dataset & LLMs & Prompting & Success (\%) & Weighted-F1 (\%) & Runtime (s) & Consistency (\%) \\ \midrule
\multirow{12}{*}{CGSS} & \multirow{6}{*}{Llama3-8B} & I/O Prompt (ICLR 2022) & 22.22 & 22.09 & 3.39 & 100 \\
 &  & Fine Tuning (ICLR 2022) & 41.41 & 41.12 & 0.31 & 100 \\
 &  & CoT (NeurIPS 2022) & 52.94 & 55.16 & 3.07 & 98 \\
 &  & SC-CoT (ICLR 2023) & 64.00 & 69.85 & 12.99 & 100 \\
 &  & ToT (NeurIPS 2023) & 66.59 & 68.01 & 12.11 & 100 \\
 &  & RFoT (ours) & 78.43 & 80.52 & 15.46 & 100 \\ \cmidrule(r){2-7}
 & \multirow{6}{*}{Qwen2.5-7B} & I/O Prompt (ICLR 2022) & 19.00 & 21.97 & 11.08 & 75 \\
 &  & Fine Tuning (ICLR 2022) & 40.00 & 34.85 & 11.25 & 100 \\
 &  & CoT (NeurIPS 2022) & 35.29 & 43.09 & 6.40 & 98 \\
 &  & SC-CoT (ICLR 2023) & 41.00 & 43.72 & 27.85 & 100 \\
 &  & ToT (NeurIPS 2023) & 43.19 & 43.10 & 22.09 & 98 \\
 &  & RFoT (ours) & 47.06 & 54.26 & 13.48 & 100 \\ \midrule
 \multirow{12}{*}{ESS} & \multirow{6}{*}{Llama3-8B} & I/O Prompt (ICLR 2022) & 35.71 & 32.14 & 3.52 & 100 \\
 &  & Fine Tuning (ICLR 2022) & 63.26 & 65.20 & 0.32 & 97 \\
 &  & CoT (NeurIPS 2022) & 49.02 & 51.37 & 3.14 & 86 \\
 &  & SC-CoT (ICLR 2023) & 60.00 & 58.35 & 9.68 & 100 \\
 &  & ToT (NeurIPS 2023) & 62.11 & 59.96 & 10.09 & 100 \\
 &  & RFoT (ours) & 68.63 & 68.51 & 12.56 & 74 \\ \cmidrule(r){2-7}
 & \multirow{6}{*}{Qwen2.5-7B} & I/O Prompt (ICLR 2022) & 20.00 & 20.98 & 11.02 & 68 \\
 &  & Fine Tuning (ICLR 2022) & 61.00 & 60.10 & 11.13 & 100 \\
 &  & CoT (NeurIPS 2022) & 54.90 & 54.42 & 2.34 & 92 \\
 &  & SC-CoT (ICLR 2023) & 57.00 & 57.48 & 24.62 & 100 \\
 &  & ToT (NeurIPS 2023) & 63.79 & 60.81 & 22.09 & 100 \\
 &  & RFoT (ours) & 72.55 & 72.12 & 12.75 & 78 \\ \bottomrule
\end{tabular} 
\end{table*}

\subsubsection{Happiness Prediction}
Happiness prediction is a common questionnaire topic in computational social science \cite{ijcai2022_Lilin,TCSS_2025}, where the goal is to analyze the happiness level of respondents via answering a questionnaire designed by sociologists from different aspects. Specifically, given a questionnaire \( x^{(n)} = \{ x^{(n)}_1, x^{(n)}_2, \dots, x^{(n)}_j \} \), where \( x^{(n)} \in \mathcal{X} \) and \( x^{(n)}_j \) represents the \( j \)-th question-answer pair \( x^{(n)} \), the goal is to predict the happiness level \( y_{ic} \in \mathcal{Y} \) from the candidate levels \( \mathcal{Y} \).

\paragraph{Happiness Dataset} For considering mental health from different perspectives and different cultures, the two happiness prediction datasets are utilized in this work as an example. We conduct comprehensive experiments to evaluate the performance of our proposed RFoT prompting on the public Chinese General Social Survey (CGSS), and the European Social Survey (ESS) datasets.

\begin{itemize}[leftmargin=*]

    \item \textbf{Chinese General Social Survey (CGSS)\footnote{http://cgss.ruc.edu.cn/}.} CGSS is an open shared and large-scale social online investigation dataset, with the subject being Chinese families containing more than 124 questions per sample. It aims to systematically monitor the changing relationship between social structure and quality of life in both urban and rural China. Social structure refers to dimensions of social group and organization as well as networks of social relationships. Quality of life is the objective and subjective aspects of people's well-being both at the individual and aggregate levels.

    \item \textbf{European Social Survey (ESS)\footnote{https://ess-search.nsd.no/}} The ESS is an academically driven and multi-country online survey that includes 50,000 samples and 102 questions per one over 38 countries to date. It has been extensively used to effectively assess the progress of nations and develop a series of European social indicators around citizens' happiness and well-being. Many studies in sociology, economics, and even politics have utilized ESS data. Note that happiness is scaled by 5 levels from very unhappy to very happy, where a higher level indicates more happiness. 

\end{itemize}

 \begin{table}[t] \small
 	\centering
 	\caption{The statistics of CGSS and ESS questionnaire datasets.}
 	\label{dataset_descirbe}
 	\begin{tabular}{ccccc}
 		\toprule
 		\multirow{2}{*}{\makecell{Mental \\ State}} & \multicolumn{2}{c}{CGSS} & \multicolumn{2}{c}{ESS}\\
 		\cmidrule(r){2-3}		\cmidrule(r){4-5}
 		& Sample & Turn & Sample & Turn \\
 		\midrule
 		Very unhappy & 77 & 124 & 721 & 102  \\
 		Unhappy & 315 & 124 & 1515 & 102 \\
 		Neutral & 630 & 124 & 3991 & 102 \\
 		Happy & 1743 & 124 & 6453 & 102 \\
 		Very happy & 422 & 124 & 2877 & 102 \\ \bottomrule
 	\end{tabular}
 \end{table}

The statistics of these three datasets after pre-processing are presented in Table~\ref{dataset_descirbe}. 

\begin{figure*}[ht]
    \centering
    \includegraphics[width=1\linewidth]{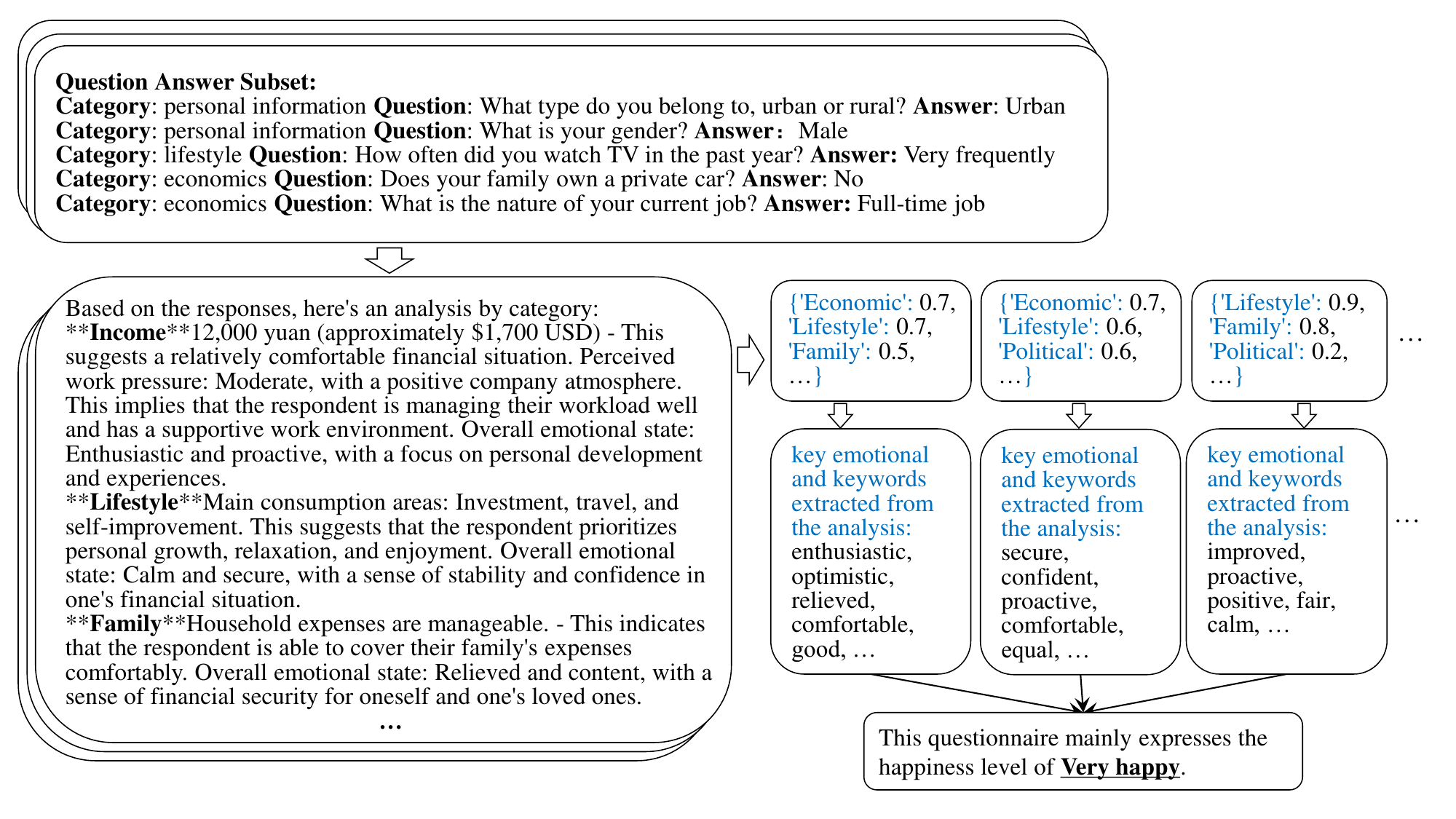}
    \caption{The case study on happiness prediction. }
    \label{fig:case_study_happiness}
\end{figure*}

\paragraph{Results}

As shown in Table \ref{tab:LLMs_Happiness_results}, the reasoning results by Llama3-8B and Qwen2.5-7B LLMs on two datasets demonstrate that our RFoT has significant improvements over all the baselines covered mainstream LLMs prompting methods. Specifically, the I/O prompting without intermediate steps (i.e., thoughts) performs more poorly than other X of thought prompting with some intermediate steps. It indicates the benefits of X of thought prompting for LLMs, which highly aligns with previous works \cite{Chain_of_Thought_2022,Tree_of_thoughts_2023,Self_Consistency_COT_2023}. Surprisingly enough, the fine-tuning method performs better than some X of thought prompting such as I/O prompting, CoT, and SC-CoT on CGSS and ESS datasets. It indicates that the X of thought can not work well in a specific area such as social science problems rather than the general task. 
Our proposed RFoT is better than other baselines in two datasets and two LLMs, which due to the RFoT can explore more combinations of thoughts generated by LLMs and well fitted in the survey analysis problems as illustrated in Fig.~\ref{fig:Random_Questionnaire}.

On the other hand, more steps will result in more average time-consuming across almost all the datasets and LLMs. Note that the SC-CoT usually has the highest time complexity because of the multiple times LLM running even though the accuracy improvements. For the metric of consistency, the results indicate the limitations of generation language models represented by LLMs, which is one of the most important future works in this area.

\paragraph{Case Study}
As shown in Fig.~\ref{fig:case_study_happiness}, the case study indicates the effectiveness of our proposed RFoT. The question-answer pairs in a questionnaire are randomly selected from different categories and are combined in a new questionnaire, which benefits focusing on main factors with various categories and excluding some noises. The question-answer analysis is generated by an LLM from different categories, such as personal information, lifestyle, and economics. Based on this, the category importance and its keywords related to the happiness state are calculated by Shapley value. Then the final prediction results can be voted on and generated from different emotional triggers and categories.

\section{Conclusions and Limitations}

\paragraph{Conclusions} The random forest of thoughts prompting provides a new way to explore a rich result space and enhance the reasoning performance in specific fields for contemporary LMs. At the same time, a post-hoc explanation method is integrated into RFoT for more comprehensive and trustworthy feature selection, providing a way to solve complex problems in a specific area that are not easily formalized, such as computational social science. 
\paragraph{Limitations} The few-shot RRoT requires predefined samples to prompt LLMs in a specific problem. In response, the representative samples selected from a dataset based on a sample-based explanation will be a significant method. Also, because of the effectiveness of the computation, it is very time-consuming to calculate the contribution of words and categories for selecting the top thoughts. In the future, we aim to optimize these two issues for better reasoning performance of LLMs in some specific areas.


\bibliography{RFoT_main}




\end{document}